\documentclass[runningheads]{llncs}
\usepackage[T1]{fontenc}
\usepackage{fancyvrb} 
\usepackage{multicol} 
\usepackage{graphicx}
\usepackage{booktabs} 

\begin{document}
\title{Evaluating LLM Translation for Prompt-Enhanced ACT-R and Soar Models}
%
%
\author{Parth Ganeriwala\inst{1} \and
Siyu Wu\inst{2} \and\\ Siddhartha Bhattacharyya\inst{1} \and Frank E. Ritter\inst{2}}
\authorrunning{Ganeriwala et al.}
%
\institute{College of COES, Florida Institute of Technology, Melbourne, FL \and
College of IST, The Pennsylvania State University, Centre County, PA\\
\email{\{pganeriwala2022,sbhattacharyya\}@fit.edu},\\
\email{\{sfw5621,frank.ritter\}@psu.edu}}
\maketitle              
\begin{abstract}
Translating cognitive models across heterogeneous cognitive architectures is a critical yet unresolved challenge in computational cognitive science, with significant implications for model reusability, interoperability, and cross-framework validation. This paper explores the feasibility of using a Large Language Model (LLM) to facilitate automated translation between ACT-R and Soar, two widely adopted cognitive modeling frameworks. We explore the feasibility of LLM-assisted translation and identify translation patterns, offering a methodological foundation for future quantitative evaluation. Through a series of controlled translation experiments and iterative prompt engineering cycles, we identify and validate a generalized translation pattern that supports bi-directional model interoperability. Our findings reveal that while LLMs demonstrate promising capabilities in syntactic scaffolding and partial semantic mapping, human-in-the-loop intervention remains critical for ensuring executable and cognitively plausible model outputs. The study offers insights into the potentials and limitations of leveraging LLMs for cross-architecture translation and provides a methodological foundation for future research on automated cognitive model transformation and hybrid architecture integration.


\end{abstract}

\section{Introduction}

The exploration of human cognition and decision-making processes has been a longstanding pursuit in cognitive science and artificial intelligence research \cite{gluck2006modeling,wu2024understanding,ho2022cognitive}. Two prominent frameworks for cognitive modeling are ACT-R \cite{anderson2009can} and Soar \cite{laird2019soar}. These frameworks serve as robust tools for simulating human behavior across various cognitive tasks. However, the traditional process of developing models within the ACT-R and Soar entails complex coding, thus delaying widespread adoption and accessibility. 

The translation between cognitive architectures—specifically from Soar to ACT-R and vice versa—represents a challenging yet crucial endeavor in cognitive science and AI research as each possesses distinct strengths, representational approaches, and architectural designs. As a result, interoperability between these frameworks has been limited. The need for robust translation mechanisms arises from their foundational differences. ACT-R models cognition through buffer manipulation and rule-based procedural execution within a modular system. In contrast, Soar emphasizes operator-based control hierarchies within a unified working memory and decision cycle. These divergent paradigms yield cognitive models that are structurally and semantically incompatible without significant adaptation. Importantly, translation across these frameworks is not merely a syntactic transformation; it requires a conceptual mapping of cognitive processes, memory structures, and control flow mechanisms. Enabling such interoperability supports comparative model evaluation, replication of findings across architectures, and reuse of cognitive task logic in hybrid or evolving frameworks. Prior efforts have attempted to implement the same cognitive model across both ACT-R and Soar to study architectural correspondences and differences, highlighting the challenges in translation fidelity and semantic alignment \cite{jones2007comparing,muller2008implementing}.

This study emphasizes LLM-assisted translation to reduce the manual burden of porting models across architectures and to leverage natural language understanding in formal model development workflows. Recent advancements in Large Language Models (LLMs), such as ChatGPT developed by OpenAI, offer promising avenues for facilitating this translation process. ChatGPT demonstrates remarkable proficiency in understanding and generating human-like language across diverse contexts and tasks. Harnessing its capabilities for translating between Soar and ACT-R not only bridges the gap between these cognitive architectures but also opens new opportunities for novel applications and insights in cognitive modeling research. Recent studies have explored how LLMs can be integrated with cognitive architectures or leveraged to augment symbolic reasoning tasks, suggesting a growing interest in hybrid cognitive-symbolic systems \cite{romero2023synergistic,joshi2023augmenting}.

We conduct a systematic evaluation of LLM-assisted translation across representative cognitive tasks, focusing on the transformation of declarative knowledge structures and production rules. Our methodology combines prompt engineering with human-in-the-loop refinement to assess translation fidelity, identify recurrent failure patterns, and establish prompt strategies that improve reliability. These findings offer new insights into the role of LLMs in supporting cognitive model development and lay the groundwork for future work in cross-architecture interoperability.

\section{Related Work}

\subsection{A Common Model of Cognition}
The Common Model of Cognition (CMC) embodies a unified theory of cognition \cite{Newell1994,laird2017standard}, a theoretical framework that presents a model of human cognition codified as a computational architecture. The CMC is a brain-inspired framework validated by large-scale neuroscience data. The CMC identifies core components and processes fundamental to human cognition, including memory, perception, motor actions, and decision-making. The model assumes a cyclical process where these components interact to produce human behavior. The CMC includes a feature-based declarative long-term memory, a buffer-based working memory, a system for pattern-directed action invocation stored in procedural memory, and specialized systems for perception and action \cite{stocco2021analysis}.

The CMC integrates essential features from various cognitive architectures \cite{laird2019soar,anderson2009can,kotserubaT25}, which propose a set of fixed mechanisms to model human behavior, functioning akin to agents and aiming for a unified representation of the mind. By using task-specific knowledge, these architectures not only simulate but also explain behavior through direct examination and real-time reasoning tracing. Two representative cognitive architectures related to the CMC are ACT-R and Soar \cite{laird2017standard}, which have been commonly used for modeling human cognition and decision-making processes.

\subsection{Cognitive Architectures: ACT-R and Soar}
ACT-R is a cognitive architecture and a theory of simulating and understanding human cognition\cite{anderson2019,ACTR}. Its theory is embodied in the ACT-R software, through which we can construct models that can store, retrieve, and process knowledge, as well as explain and predict performance\cite{Bothell2017}.
There are currently two kinds of knowledge representations in ACT-R, and they are declarative knowledge and procedural knowledge. Declarative knowledge consists of chunks of memory (e.g., apple is a kind of fruit), while procedural knowledge performs basic operations, moves data among buffers, and identifies the next instructions to be executed (e.g., to submit your answer, you have to click submit bottom). When the model is driving a bus in first-person perspective, these pieces of information will contain information such as what visual items are presented to look at and what tasks to do next.

Soar is a general cognitive architecture that provides a computational infrastructure that resembles the cognitive capabilities exhibited by a human. Soar implements knowledge-intensive reasoning that enables execution of rules based on the context. It also has the capability to integrate learning into the intelligent agent using chunking or reinforcement learning. Soar has its origins in the groundbreaking work done by Newell and Simon around the 1950s through the mid-1970s, also inspired by the ”General Problem Solver” created by George Ernst and Newell. While ACT-R was designed to model human behavior, Soar was inspired by it. Current understanding and hypotheses regarding cognitive architecture are incorporated into Soar 9, which has been in development for over 30 years and continues to evolve gradually. Soar’s general computing concept is based on: objectives, problem spaces, states and operators \cite{laird2019soar,Newell1994}. Soar encompasses multiple memory constructs (e.g., semantic, episodic, etc.) and learning mechanisms (e.g., reinforcement, chunking etc.) and is a programmable architecture with an embedded theory. This enables executing Soar models on embedded system platforms and studying the design problem through rapid prototyping and simulation.

\section{A Case Study on ACT-R and Soar Translation} \label{sec:case-study}

To evaluate the feasibility of LLM-assisted translation between cognitive architectures, we developed functionally equivalent counting models in both ACT-R and Soar. Each model implemented production rules to initialize, increment, and terminate a simple counting task \cite{laird2019soar,Bothell2017}. These models served as controlled translation targets, enabling focused assessment of structural fidelity and semantic alignment in translated outputs. Our goal was to generate the Soar model from its ACT-R counterpart and vice versa, using ChatGPT as the translation engine.

\begin{figure*}[h!tbp]
\centering
\includegraphics[width=\linewidth]{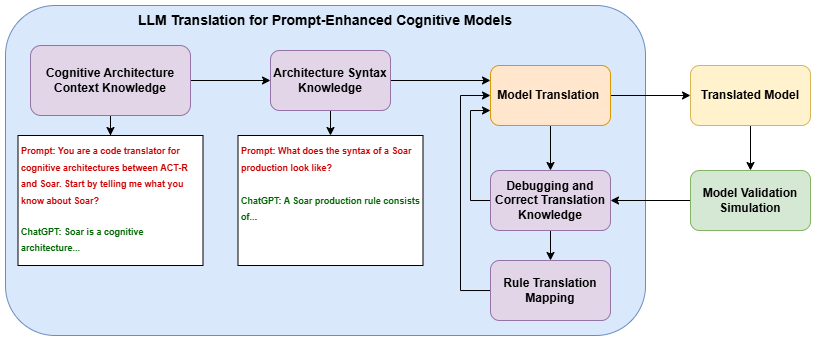}
\caption{Iterative human-in-the-loop methodology for translating cognitive models between ACT-R and Soar using ChatGPT.}
\label{fig:methodology}
\end{figure*}

Our translation framework (Figure~\ref{fig:methodology}) follows a structured pipeline involving prompt initialization, model generation, human-in-the-loop (HITL) correction, and simulation. ChatGPT is primed with architectural and syntactic knowledge before producing translated models. Human reviewers refine these outputs iteratively until the result satisfies syntactic validity and behavioral equivalence. Full translation examples and prompt patterns are discussed in Sections~\ref{sec:translation-results} and~\ref{sec:translation-patterns}.

Each translation experiment was structured into clearly defined phases to ensure consistency and reproducibility. In the initialization phase, prompts were carefully constructed to encode both the semantic and syntactic specifications of the source model. These prompts described the cognitive task, the architecture-specific control flow, and the expected memory representations. For ACT-R, prompts included detailed descriptions of buffer structures and chunk types. For Soar, prompts described working memory elements (WMEs), operator proposal-application logic, and substate control structures.

ChatGPT (v3.5 and 4o) was then queried to generate the translated model. As expected, the general-purpose nature of the LLM led to early-stage outputs that exhibited structural inconsistencies and semantic omissions. Common errors included incorrect buffer-slot mappings in ACT-R, misassigned WMEs in Soar, and invalid or fragmented rule conditions. These outputs were reviewed and we provided targeted feedback, diagnosing model violations, clarifying misunderstood constructs, and suggesting domain-consistent representations. This feedback was integrated into subsequent prompt revisions, forming the basis of the iterative HITL loop. Refinement continued until each translated model passed expert verification for syntactic correctness and cognitive equivalence. All iterations, modifications, and expert interventions were logged to facilitate traceability, meta-analysis, and the identification of generalizable translation strategies.
Expert verification in our study was defined as the combined assessment of (a) syntactic correctness (whether the model compiled and ran in the target architecture), (b) semantic fidelity (whether task logic and rule conditions were preserved), and (c) execution validity (whether the translated model produced expected outputs such as correct counts). These criteria guided acceptance or rejection of each translation stage.
\begin{figure}[htbp]
\centering
\begin{minipage}[t]{0.48\linewidth}
\centering
\textbf{Soar Translation (Excerpt)}
\begin{Verbatim}[fontsize=\footnotesize]
sp {apply*initialize-count
  (state <s> ^operator <o>)
  (<o> ^name initialize-count)
  -->
  (<s> ^count 1 ^end 6)}
\end{Verbatim}
\end{minipage}
\hfill
\begin{minipage}[t]{0.48\linewidth}
\centering
\textbf{ACT-R Translation (Excerpt)}
\begin{Verbatim}[fontsize=\footnotesize]
(p initialize-count
   =goal>
     ISA count-goal
     count nil
     end =end
   ==>
   =goal>
     count 1
   !output! (print-number 1))
\end{Verbatim}
\end{minipage}
\vspace{1mm}
\caption{Representative production rules from the LLM-generated translations of a counting task in Soar and ACT-R. Full model listings are available in our project repository at \url{https://github.com/ParthGaneriwala/llm-cognitive-model-translation}.}
\label{fig:soar-actr-example}
\end{figure}

\section{Translation Results}
\label{sec:translation-results}
We conducted bidirectional translation tasks between Soar and ACT-R and evaluated their syntactic correctness, task completion behavior, and architectural alignment. Successful execution of the translated counting models (see Figure~\ref{fig:soar-actr-example}) validated that the LLM-generated outputs—when refined through human-in-the-loop correction—preserved task logic across both frameworks. 
While the evaluation was primarily qualitative, we also recorded basic metrics across the translation experiments (Table 2).
\begin{table}[h]
    \centering
    \begin{tabular}{l|c|c|c|c}
        \textbf{Direction} & \textbf{\# Attempts} & \textbf{Avg. Iter} & \textbf{Init Syntax Err} & \textbf{Success After Iter. 2} \\
        \hline
        ACT-R $\rightarrow$ Soar & 6 & 2.5 & 67\% & 100\% \\
        Soar $\rightarrow$ ACT-R & 6 & 2.2 & 72\% & 100\% \\
    \end{tabular}
    \caption{Quantitative summary of translation attempts showing the number of attempts, average prompt-correction iterations, percentage of initial syntax errors, and success rate after two iterations.}
    \label{tab:translation-summary}
\end{table}
These results demonstrate that while initial LLM outputs are error-prone, structured prompts and HITL corrections reliably converge to executable models.

The \textit{translation from Soar to ACT-R} posed a unique set of representational and structural challenges. Soar’s cognitive control is governed by the dynamic creation of problem spaces, substate formation, and operator-based elaboration, all of which are orchestrated through a cyclical decision process. Mapping these procedural dynamics into ACT-R required conceptual flattening of Soar’s hierarchical state management into ACT-R’s localized buffer-centric model of cognition. In particular, Soar operators—often defined through interdependent propose and apply rules with implicit working memory manipulations—had to be decomposed into ACT-R productions with explicitly stated preconditions and buffer actions. Soar’s working memory elements (WMEs), structured as identifier–attribute–value triples, were converted into ACT-R chunks that conform to a fixed chunk-type specification. This transformation required flattening deeply nested WMEs and re-encoding their semantics to match ACT-R’s declarative memory representation. Additionally, ACT-R’s production rules necessitate precise control over the flow of information between goal and retrieval buffers, which demanded the explicit encoding of operations that are otherwise handled implicitly in Soar through preference resolutions and substate activations.

Human oversight played a critical role in ensuring the cognitive and operational fidelity of the translated ACT-R models. We verified that transitions and task logic embedded in Soar were properly restructured into production sequences and chunk updates in ACT-R. Furthermore, implicit control structures in Soar—such as impasse resolution, subgoal formation, and termination conditions—had to be explicitly represented and tracked in ACT-R, often requiring the introduction of additional productions or buffer manipulations.

The \textit{reverse translation, from ACT-R to Soar}, involved a different set of technical challenges. ACT-R’s rule-based interaction across modular buffers (e.g., goal, retrieval, visual) had to be reinterpreted in terms of Soar’s unified working memory and operator-driven decision-making. Declarative memory chunks in ACT-R, typically accessed through retrieval buffers and matched using slot-value constraints, were translated into WMEs with explicit graph-like relationships in Soar. In addition, ACT-R’s production rules, which encode procedural knowledge using a reactive stimulus-response model, lacked a one-to-one structural analogue in Soar’s elaboration-operator structure. Consequently, high-level ACT-R rules often had to be decomposed into sequential Soar operators that explicitly track intermediate state transitions and control flow.

Across both directions, the iterative refinement of LLM-generated code was essential. ChatGPT’s initial outputs frequently exhibited syntactic violations, architectural inconsistencies, or semantically ambiguous constructs. The human-in-the-loop (HITL) correction cycle enabled the systematic revision of prompts, reinforcement of architectural constraints, and guided transformation of ambiguous outputs into executable models. This iterative feedback loop was instrumental not only in correcting structural deficiencies but also in distilling robust prompt patterns for future translation tasks.

\section{Translation Patterns}
\label{sec:translation-patterns}
Analysis of the translated models revealed a series of systematic translation patterns applicable across both Soar-to-ACT-R and ACT-R-to-Soar conversions. These patterns emerged through iterative prompt execution cycles and were reinforced through expert interventions and model verification. The results highlight how large language models, when scaffolded appropriately, can facilitate meaningful transformation of symbolic cognitive systems across architectural boundaries.

\subsection{Structural and Semantic Transformation}

One of the most prevalent translation strategies involved the structural flattening and semantic decomposition of rules and control logic. In the Soar-to-ACT-R direction, hierarchical operator structures and nested propose–apply cycles were decomposed into localized ACT-R productions with clearly defined preconditions and buffer-manipulating actions. This flattening was essential for aligning Soar’s state-anchored, graph-structured control representations with ACT-R’s modular buffer architecture. Similarly, implicit control flow constructs in Soar—such as substate management, operator preference evaluation, and automatic impasse resolution—had to be made explicit in ACT-R. This often required introducing additional productions or manually encoding transitions that Soar would otherwise handle implicitly.

In the reverse direction, ACT-R’s concurrent buffer access and slot-specific reasoning required decomposition into Soar’s operator-centric processing model. ACT-R rules that simultaneously matched conditions in multiple buffers were split into sequences of Soar productions, each tied to an operator propose–apply cycle. To preserve execution fidelity, temporal coordination and rule triggering logic had to be carefully distributed across the Soar model’s elaboration and application stages.

\subsection{Knowledge Representation Mapping}

Translating models across architectures necessitated substantial reformatting of knowledge structures to fit the semantic and syntactic expectations of the target system. Soar’s working memory elements (WMEs), represented as flexible attribute–value graphs anchored in a global state identifier, were transformed into ACT-R chunks that follow typed schema declarations with fixed slot-value pairs. This process involved flattening multi-level WMEs, reassigning variable names, and introducing new chunk-types to capture intermediate state distinctions.

Conversely, ACT-R’s declarative memory representations—comprising named chunks stored and retrieved through buffer access—were mapped to Soar’s untyped and more flexible WME graph structure. This required re-encoding slot-value relationships using attribute triples, and in many cases, maintaining identity across model states via consistent variable binding. Maintaining the logical equivalence of relational mappings during this transformation process was nontrivial. In particular, references to previously stored chunks in ACT-R needed to be carefully reconstructed in Soar using nested WMEs and symbolic references. The structural integrity and referential cohesion of task-critical knowledge required detailed expert inspection and iterative reformulation.

\subsection{Prompt Engineering and Expert Refinement}

The effectiveness of LLM-assisted translation was strongly influenced by the structure and granularity of prompts. Prompts that supplied layered architectural context—including definitions of memory systems, rule syntax, control flow, and task framing—significantly improved output quality. Decomposing translation tasks into sequential subprompts (e.g., defining chunk types, then generating rule scaffolds, then simulating behavior) enabled clearer cognitive mappings and reduced syntactic errors. These findings are consistent with broader research on prompt engineering and human-in-the-loop (HITL) frameworks for LLMs, which emphasize the role of task-specific prompts and iterative refinement in guiding reliable model behavior \cite{Sahoo2024,mosqueira2023human}.

Explicit error feedback also played a crucial role. When initial generations produced invalid or incomplete outputs, providing the LLM with direct debugging signals—such as “missing slot assignment,” “incorrect buffer reference,” or “invalid syntax for operator application”—often led to materially improved results on subsequent iterations. This pattern reinforces the LLM’s ability to refine outputs when guided by corrective human intent.

\begin{table}[h!]
\centering
\caption{Prompt types used during translation experiments}
\label{tab:prompt-types}
\begin{tabular}{p{3.2cm} p{5.2cm} p{4.1cm}}
\toprule
\textbf{Prompt Type} & \textbf{Purpose} & \textbf{Example Pattern} \\
\midrule
Initialization Prompt & Establish role, task scope, and architectural context & \texttt{You are a code translator between Soar and ACT-R...} \\
Declarative Structure Prompt & Define memory schema and chunk types or WMEs for a task & \texttt{Define chunk-types for representing count-order and goal.} \\
Procedural Rule Prompt & Generate a rule in the target architecture from a source rule & \texttt{Translate this Soar propose rule into an ACT-R production.} \\
Sequencing Prompt & Scaffold multi-step outputs for complex tasks & \texttt{Step 1: define chunk types. Step 2: generate initial production.} \\
Debugging Feedback Prompt & Correct syntax or semantic errors in prior output & \texttt{You forgot the buffer condition in the goal module. Fix that.} \\
Post-execution Prompt & Adjust model based on execution outcome or trace errors & \texttt{The model fails when count reaches 6. Add a termination rule.} \\
\bottomrule
\end{tabular}
\end{table}

Human-in-the-loop (HITL) review proved essential across all translation directions. Initial translations frequently exhibited architectural violations such as goal state mismanagement, invalid chunk references, or malformed working memory assertions. Domain experts intervened to correct logic, rephrase prompts, and align generated structures with the expectations of the target cognitive architecture. These corrections were incorporated into prompt refinements, resulting in the gradual development of reusable prompt templates. The HITL pattern functioned not only as a safeguard but also as an iterative calibration mechanism for the LLM's internal representation of cognitive modeling principles. To demonstrate how prompt structure influenced translation outcomes, we include representative prompts from the Soar-to-ACT-R translation direction. As an example, a typical initialization prompt reads:
\begin{quote}
\small
\texttt{You are a code translator between Soar and ACT-R...}
\end{quote}
Additional prompt types and usage scenarios are summarized in Table~\ref{tab:prompt-types}. These prompt forms were critical in aligning model outputs with valid syntactic structures and achieving semantic equivalence. Debugging prompts in particular played a key role in triggering self-correction within ChatGPT’s generative loop.

\section{Conclusion}

This study explored the feasibility of translating cognitive models between ACT-R and Soar using large language models. We found that, although initial outputs often required significant correction, structured prompt engineering and human-in-the-loop refinement enabled successful, semantically faithful translations. The identification of recurring translation patterns—such as flattening, explicit memory reformulation, and syntax re-alignment—provides a basis for developing more scalable and reliable translation workflows. While the HITL framework enables reliable translation, it requires expert-level review to correct architectural violations. The general-purpose nature of ChatGPT introduces brittleness in unfamiliar task domains, and the current framework is constrained to well-scoped models such as counting. Scaling to multi-module or perceptual-motor tasks remains an open challenge. We also observed occasional hallucinations (e.g., non-existent buffer names), which were systematically identified during expert review and corrected through prompt refinement. Future work will explore automated correction strategies, hallucination detection, richer cognitive tasks, and quantitative benchmarks for accuracy and human effort reduction.
\bibliographystyle{splncs04}
\bibliography{refs}
\end{document}